\documentclass[journal]{IEEEtran}
\usepackage{multirow}
\usepackage{graphicx}
\usepackage[normalem]{ulem}
\usepackage{caption}
\usepackage{subcaption}
\useunder{\uline}{\ul}{}

\makeatletter
\def\subsubsection{%
  \@startsection
    {subsubsection}                 
    {3}                             
    {\parindent}                    
    {3.5ex plus 1.5ex minus 1.5ex}  
    {0.7ex plus .5ex minus 0ex}     
    {\normalfont\normalsize\itshape}
}
\makeatother

\usepackage{amsmath}
\usepackage{amssymb}
\usepackage[table,xcdraw]{xcolor}

\usepackage[style=ieee,sorting=ynt]{biblatex}
\ifCLASSINFOpdf
\else
\fi
\usepackage{graphicx}

\begin{document}
%
\title{Learning Through Guidance: Knowledge Distillation for Endoscopic Image Classification}
%
%
%

\author{Harshala~Gammulle*,~\IEEEmembership{Member,~IEEE,}
        Yubo~Chen*,
        Sridha~Sridharan,~\IEEEmembership{Life~Senior~Member,~IEEE}
        Travis~Klein
        and Clinton~Fookes,~\IEEEmembership{Senior~Member,~IEEE.}
\thanks{H. Gammulle, S. Sridharan, and C. Fookes are with the Signal Processing, Artificial Intelligence and Vision Technologies (SAIVT), Queensland University of Technology (QUT), Australia.}
\thanks{Y. Chen is with the School of Mechanical, Medical, and Process Engineering, QUT, Australia.}
\thanks{T. Klein is with the QUT Centre for Biomedical Technologies, QUT, Australia.}
\thanks{* H. Gammulle and Y. Chen contributed equally to this work.}}
\maketitle

\begin{abstract}
    Endoscopy plays a major role in identifying any underlying abnormalities within the gastrointestinal (GI) tract. There are multiple GI tract diseases that are life-threatening, such as precancerous lesions and other intestinal cancers. In the usual process, a diagnosis is made by a medical expert which can be prone to human errors and the accuracy of the test is also entirely dependent on the expert's level of experience. To assist medical experts in accurate diagnosis, we propose an AI-based model to automate the abnormality detection process for endoscopy data. Deep learning, specifically Convolution Neural Networks (CNNs) which are designed to perform automatic feature learning without any prior feature engineering, has recently reported great benefits for GI endoscopy image analysis. Previous research has developed models that focus only on improving performance (such as accuracy), as such, the majority of introduced models contain complex deep network architectures with a large number of parameters that require longer training times. However, there is a lack of focus on developing lightweight models which can run in low-resource environments, which are typically encountered in medical clinics. In this study, we investigate Knowledge Distillation (KD) techniques to develop an efficient GI tract endoscopy image classification method that is both lightweight and achieves satisfactory classification performance. We investigate three KD-based learning frameworks, response-based, feature-based, and relation-based mechanisms, and introduce a novel multi-head attention-based feature fusion mechanism to support relation-based learning. Compared to the existing relation-based methods that follow simplistic aggregation techniques of multi-teacher response/feature-based knowledge, we adopt the multi-head attention technique to provide flexibility towards localising and transferring important details from each teacher to better guide the student. We perform extensive evaluations on two widely used public datasets, KVASIR-V2 and Hyper-KVASIR, and our experimental results signify the merits of our proposed relation-based framework in achieving an improved lightweight model (only 51.8k trainable parameters) that can run in a resource-limited environment.   
\end{abstract}

\begin{IEEEkeywords}
Knowledge Distillation, Computer Aided Gastrointestinal Disease Detection, Medical Image Processing, Deep Learning, Artificial Intelligence.
\end{IEEEkeywords}

%
\IEEEpeerreviewmaketitle

\section{Introduction}
%
%
%
%
\IEEEPARstart{E}{xamination} of the GastroIntestinal (GI) tract through endoscopy is a vital medical procedure that plays a significant role in promoting human health and well-being. It has contributed to extending human life and improving the quality of life by identifying any underlying anomalies within the GI tract \cite{kim_gastrointestinal_2017} such as life-threatening precancerous lesions \cite{soetikno_endoscopic_2016}. GI tract examination is a powerful tool to aid medical practitioners in diagnosing and subsequently assisting patients in mitigating the likelihood of contracting diverse ailments \cite{liedlgruber_computer-aided_2011}. The implementation of an automated detection system can assist clinicians in the diagnosis and can alleviate the workload of gastroenterology specialists during GI tract examinations, enabling them to focus on critical cases. According to the findings in \cite{van_rijn_polyp_2006}, during manual screening 26\% of neoplastic polyps are missed in tandem colonoscopy. Especially, in the case where the polyps are less than 10mm, the miss rate has significantly increased. Therefore, the use of AI can assist inexperienced medical practitioners in decision-making, which would prove particularly helpful in addressing the issue of observer variability and human error.

The majority of existing methods for endoscopy data-based classification focus heavily on achieving better classification performance with respect to accuracy, yet these often require a lot of processing power to obtain real-time responses, which is usually beyond the capabilities of a typical laptop/iPad which would be readily available in a clinical setting. However, there's a lack of research on how these well-performing heavy-weight models can be utilized to obtain efficient models that run in low-resource environments. We believe that it is an essential aspect to investigate lightweight deep models, as it leverages the capacity to be run on devices such as wireless capsule endoscopes and edge devices in hospital settings that have lower computational power. Furthermore, these lightweight models need to be reliable with accurate predictions while being suitable to run in resource-constrained environments. Our development of lightweight models as proposed in this paper will pave the way for the application of machine learning models in real-world clinical settings, which typically contain low computational resources.


Generally, most lightweight models are unable to learn as effectively as heavyweight models due to their lower number of trainable parameters that limit their ability to learn non-linear relationships within data. Therefore, achieving better performance through a lightweight deep model is quite challenging and requires careful optimal design techniques. In order to enhance the effectiveness of the lightweight models Knowledge Distillation (KD) \cite{hinton_distilling_2015}\cite{gou_knowledge_2020} techniques have been proposed. The aim of KD is to transfer knowledge learned by the heavy-weight model/s (known as the teacher) to the lightweight model (known as the student). Through this, it is expected to achieve a lightweight model which has enhanced classification ability while also able to produce efficient predictions potentially in real-time.


\begin{figure*}[htbp]
        \centering
        	\includegraphics[width=0.75\textwidth]{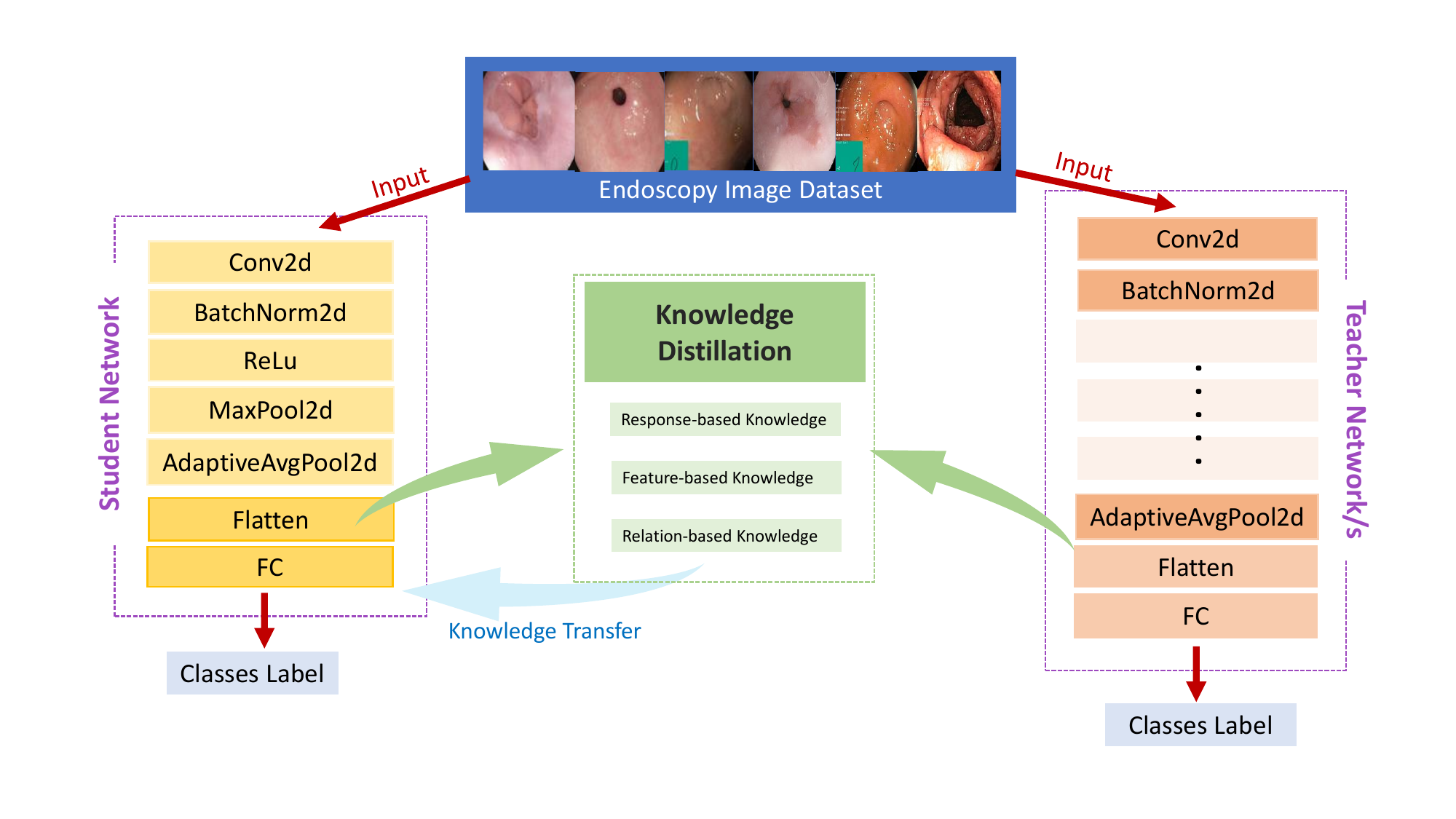}
	\caption{An overview of the Knowledge Distillation (KD) frameworks that we utilised for obtaining an enhanced lightweight endoscopic image classification model. We investigate three knowledge-based KD methods; response-based, feature-based and relation-based methods, and introduce a novel relation-based knowledge transfer mechanism by utilising multi-head attention to further enhance the performance.}
	\label{fig:framework}
\end{figure*}

We summarise our main contributions below,

\begin{itemize}
    \item We benchmark the performance of state-of-the-art ResNet models (as teachers) and our custom lightweight model (the student) on two popular gastrointestinal disease detection datasets.
    \item We propose novel algorithms for single and multi-teacher-based knowledge distillation using response-based and feature-based techniques to enhance the performance of our lightweight student.
    \item We introduce a transformer encoder block to facilitate multi-teacher-based relational knowledge distillation.
    \item We conduct extensive evaluations using two publicly available benchmarks and demonstrate the viability of our proposed knowledge distillation technique for effective and efficient knowledge transfer. 
\end{itemize}

Our study centres on investigating effective ways of performing knowledge distillation for creating lightweight models for the task of multi-class image classification, in the realm of gastrointestinal disease detection. Our main focus is on investigating lightweight models that are reliable and able to run on resource-constraint devices, as such we investigate multiple KD frameworks (illustrated in Fig. \ref{fig:framework}) discussed in \cite{gou_knowledge_2020} for endoscopy image classification. As the initial step we custom design our lightweight model and select three ResNet architectures \cite{he_deep_2015}, namely ResNet50, ResNet101, and ResNet152, that have been widely used in related image-classification methods as the teacher networks. Other than the individual evaluations on the selected endoscopy image datasets, we also investigated the three offline knowledge-based distillation techniques: response-based, feature-based, and relation-based. Through this, we expect to understand the optimal ways of transferring important cues to support the endoscopy image classification task and achieving a lightweight reliable model. We additionally contribute a relation-based method that learns to combine two-teacher-based knowledge to better guide the student model. As stated in \cite{gou_knowledge_2020} the relation-based KD is often used to map the relations among different layers or data samples of the single teacher-based knowledge transfer. The existing multi-teacher KD-based approaches have not investigated learning mechanisms to model the expertise of different teachers and distil their knowledge to better guide the student network. Our proposed relation-based technique is quite novel as we adopt multi-head attentions \cite{vaswani_attention_2017} to identify the most informative cues that the individual teacher provides and map them to a unified feature vector which is then utilised to guide the student network. The benefit of having such a relation-based approach is that it enables us to compare and contrast multiple teachers and select the important information from each teacher network using learnable attention heads to support student learning. Our final results signify the effectiveness of the knowledge transfer method which aided in further enhancing the classification performance of our lightweight model.

\section{Related Works}
    Detecting abnormalities in endoscopy data through computer vision techniques has been a focus of research in recent years. The majority of the studies in this area have centred around the analysis of colonoscopy data\cite{pogorelov_kvasir_2017}\cite{borgli_hyperkvasir_2020}, specifically with regard to the classification of GI tract-based diseases and anatomical landmarks \cite{gammulle_two-stream_2020}\cite{borgli_hyperkvasir_2020}, detection/ segmentation of polyps \cite{bernal_towards_2012}\cite{tajbakhsh_automated_2016}\cite{wang_efficient_2023}\cite{wang_multi-scale_2021}\cite{yue_benchmarking_2023}, detection of stomach lesions \cite{wu_real-time_2022}, detection of ulcerative colitis \cite{bossuyt_automatic_2020}, and classification of colorectal tumor\cite{shanmuga_sundaram_enhancement_2019}\cite{tamaki_computer-aided_2013}. The development of computer-aided detection for gastrointestinal (GI) endoscopy has been driven by a desire to address the potential concern of missed early GI cancer and disease precursors during endoscopic surveillance \cite{jha_comprehensive_2021}. Recent studies have developed and implemented models based on machine learning or artificial intelligence to classify and detect multiple gastrointestinal organs \cite{ozturk_residual_2021}.

When implementing machine learning models, the feature extraction step plays a key role in capturing important unique attributes from image data. The earliest works were based on extracting hand-crafted features such as local binary patterns (LBP) \cite{liu_hkbu_2017} and bidirectional marginal Fisher analysis (BMFA) \cite{naqvi_ensemble_2017}, then these would ideally be passed through a classical classification model such as Support Vector Machine (SVM) to obtain the desired outcomes. However, such feature extraction methods are highly dependent on the expert's knowledge and might not always capture the actual unique features that would support the final task. Due to this limitation, advanced techniques such as deep learning have been introduced to automatically learn and capture unique cues from the data to facilitate classification.

Nowadays the majority of the GI tract-based endoscopy image classification methods utilise deep learning models \cite{petscharnig_inception-like_2017}\cite{pogorelov_kvasir_2017}\cite{gammulle_two-stream_2020} and they have proven to effectively capture salient features that aid in distinguishing among various diseases. Through this automatic feature learning paradigm deep learning models are able to learn non-linear relationships between the data and the desired outputs without any feature engineering. It is also believed that such deep networks can achieve enhanced learning and generalization capabilities through higher network capacity \cite{gou_knowledge_2020}. Such large-scale deep network models have undoubtedly achieved outstanding success with acceptable and competitive performance in various domains. However, these large networks require large amounts of data to train from scratch (requires longer training times) and their complexity also makes it difficult for them to be implemented in real-time and resource-limited applications \cite{gou_knowledge_2020}. On the other hand, deep networks with lower network capacity often struggle to capture non-linearities among data similar to their larger network counterparts due to a lower amount of trainable parameters. Then the question lies in how these larger networks can facilitate the learning of smaller networks to achieve both efficiency and better classification performance.  



The concept of KD was introduced in \cite{hinton_distilling_2015} in order to facilitate smaller network (student) learning through the guidance of larger networks (teacher). This streamlined model can be then readily deployed in real-time or resource-constrained applications, offering greater convenience and ease of use. There are multiple works that have utilised KD-based learning for endoscopic data analysis \cite{garbay_distilling_2019}\cite{sivaprakasam_xp-net_2022}\cite{huang_real-time_2020}\cite{chavarrias-solano_knowledge_2022}. However, these methods were implemented in the context of segmentation and object detection in gastrointestinal tract imaging, as well as in the diagnosis of singular endoscopic anomalies, such as polyps. Considering the lack of research on KD-based multi-disease classification we perform extensive evaluations by following the KD methods stated in \cite{gou_knowledge_2020}. In \cite{gou_knowledge_2020} two main aspects of KD are discussed; knowledge types and distillation schemes. Knowledge types were further classified into three categories: response-based knowledge, feature-based knowledge, and relation-based knowledge. Distillation schemes were classified into offline distillation, online distillation, and self-distillation. Our experiments are based on different learning frameworks following the three different Knowledge types learned in an offline manner (Further details on these KD-based learning are included in Sec. \ref{sec:kd}). However, offline distillation still expects the prior training of the larger network (teachers) which requires a large amount of data and processing. As a solution for this, we adapt transfer learning techniques \cite{jha_comprehensive_2021}\cite{ozturk_residual_2021}\cite{sharif_deep_2021}\cite{gamage_gi-net_2019} to train the network to yield satisfactory results and competitive performance.  

As in \cite{gou_knowledge_2020}, the relation-based KD is mainly applied among different layers or data samples that incorporate only a single teacher network. In the case of two-teacher or multi-teacher settings, the response and feature-based knowledge are utilised through applying operations such as averaging \cite{park_feature-level_2020}. Some work \cite{yang_knowledge_2018} \cite{furlanello_born_2018} have utilised "born again" networks, where the student at step $t$ is utilised as the teacher of the student at $t+1$. Furthermore, the proposed multi-teacher approach in \cite{wu_multi-teacher_2019} integrate multi-teacher knowledge into a soft label which is the weighted average of soft probability distributions from multiple teachers. However, in this work, we utilise relation-based knowledge transfer with a multi-teacher framework. In order to facilitate this, we propose a multi-head attention-based fusion mechanism to learn and generate informative feature vectors to support the guidance of the student network. We gained the inspiration for the multi-head attention from \cite{vaswani_attention_2017} which is the main building block of Transformer networks. Even though such attention-based networks are much more capable of learning non-linearity among data through a single network, there is a lack of research that has been done in utilising such attention towards data fusion \cite{zhang_multi-head_2022}. In our study the idea of adopting such a fusion mechanism to facilitate relation-based multi-teacher knowledge transfer is novel and our final results signify the importance of our proposed novelty.

\section{Methodology and Materials}
    This section describes the methods and techniques we followed to achieve the final relation-based KD framework that contributed to enhanced knowledge transfer to the lightweight model. In order to perform the expected experiments on three KD techniques we have considered multiple network formulations and we discuss the details below.

\subsection{Knowledge Distillation for Endoscopy Image Classification}
\label{sec:kd}
        
        The proposed endoscopy image classification model is capable of detecting various abnormalities and anatomical landmarks within gastrointestinal tract-based images. Once the endoscopy image, $x$ is passed through the classification model, it outputs the class label $y$ of the corresponding abnormality or the anatomical landmark. Considering the nature of the problem we have utilised multiple CNN variants that suits the image classification task. However, as evidenced by the previous works large complex CNNs are more capable of achieving better performance with respect to accuracy. Yet the large network capacity of these models limits their application to real-world tasks that operate in resource-limited environments. Such that KD techniques aid in transferring the learned knowledge of these complex networks to a low-capacity network which then can be utilised to execute in low-resource environments (e.g. mobile phones) with enhanced performance. The proposed KD framework is illustrated in Fig. \ref{fig:framework}. Through the proposed approach we expect to enable the student model to acquire nuanced information and approximate the outputs of teacher models while simultaneously optimising computational efficiency, compactness, and generalisation performance. Furthermore, the utilisation of three types of knowledge facilitates the acquisition of knowledge by the student model through the teacher model's uncertainty, thereby enhancing its resilience to imprecise or ambiguous data. 
        
        The following sections provide details on three different KD techniques: response-based, feature-based, and relation-based.

        \begin{figure*}[htbp]
                \centering
                	\includegraphics[width=0.75\textwidth]{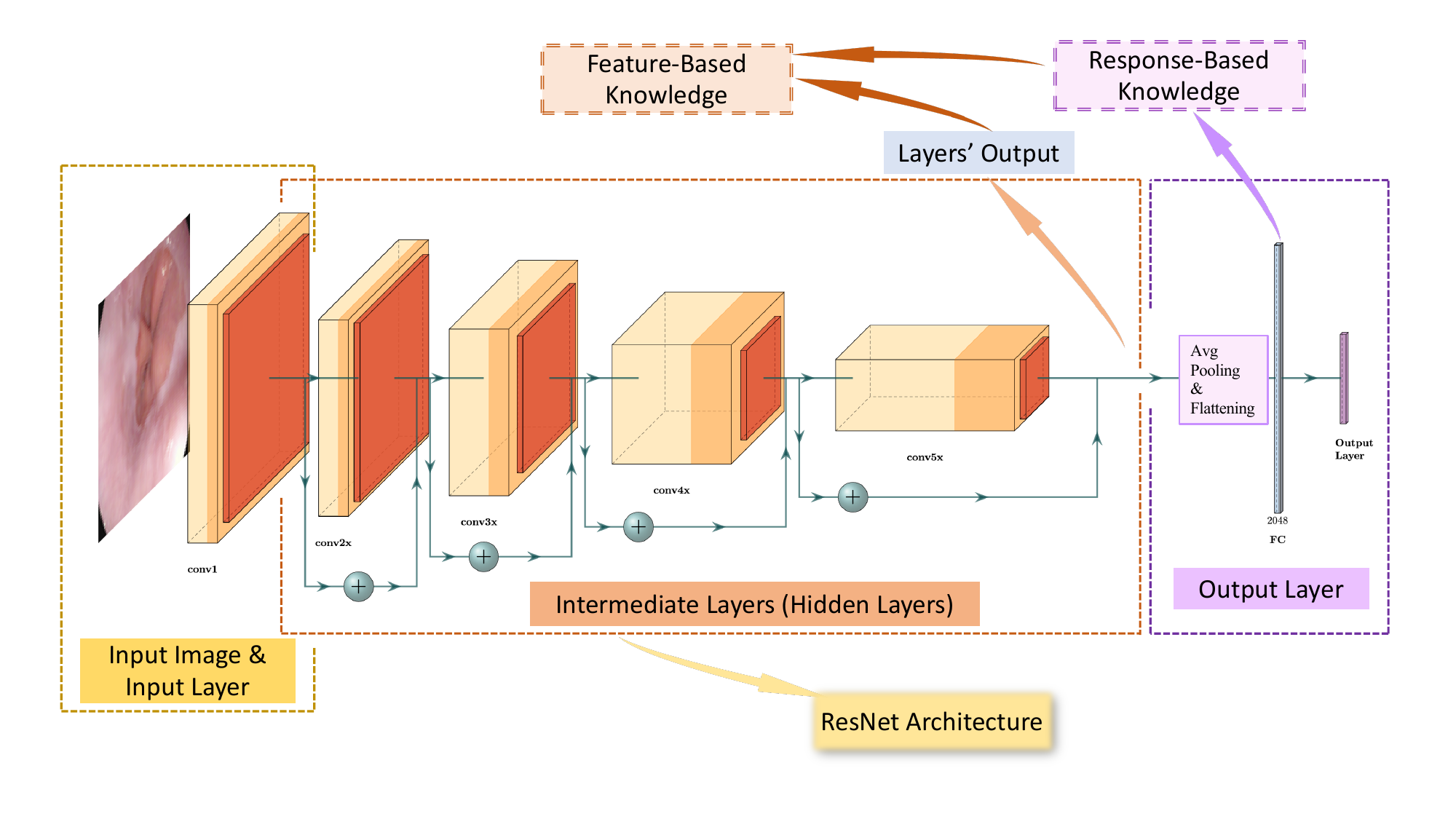}
        	\caption{The visual illustration of the origins of knowledge in a deep teacher network, including response-based knowledge and feature-based knowledge. Note that the two-teacher relation-based knowledge is not included here due to the complexity of visualising it. Please see Fig. \ref{fig:relational_kd} for the relation-based knowledge transfer mechanism that we propose in this study. }
        	\label{fig:Knowledge}
        \end{figure*}

    \subsubsection{Response-based Knowledge in Teacher Models}
        The term "response-based knowledge" typically pertains to the neural response exhibited by the final output layers of the teacher model, it is illustrated in Fig. \ref{fig:Knowledge}. The knowledge that is response-based involves replicating the final prediction of the teacher model in a direct manner. The process employed to extract response-based knowledge for our endoscopic imaging classification is commonly referred to as soft targets. The SoftMax function can be utilised to estimate the likelihood of input belonging to certain classes, which are commonly referred to as soft targets. This can be written as,
        \begin{equation}
            \sigma(z_i) = \frac{e^{z_{i}}}{\sum_{j=1}^K e^{z_{j}}} \ \ \ for\ i=1,2,\dots,K,
            \label{eq:SoftMax}
        \end{equation}
        where $z_{i}$ is the vector of logits $z$ for $i_th$ class and the logits are the outputs of the last fully connected layer of a deep neural network. The distillation loss\cite{gou_knowledge_2020} for response-based knowledge can be formulated as,
        \begin{equation}
            L_{ResD}(z_t, z_s) = \mathfrak{L}_{R}(z_t, z_s),
        \end{equation}
        where $\mathfrak{L}_{R}(z_t, z_s)$ is the divergence loss of logits, the $z_t$ and $z_s$ are teacher and student logits, respectively. Therefore, the distillation loss for soft logits can be rewritten as,
        \begin{equation}
            L_{ResD}(\sigma(z_t), \sigma(z_s)) = \mathfrak{L}_{R}(\sigma(z_t), \sigma(z_s)).
        \end{equation}
        
        The knowledge that is based on response typically depends on the final layer's output, which limits its ability to account for the intermediate-level guidance provided by the teacher model. To surmount this constraint, additional knowledge categories were incorporated to guide the student model.
        
    \subsubsection{Feature-based Knowledge in Teacher Models}
        Feature-based knowledge is derived from the outputs of both the last layer and the intermediate layers refer, as shown in Fig. \ref{fig:Knowledge}. The proposed process involves aligning the feature activations of the teacher and student models and subsequently supervising the student model using the intermediate layer outputs of the teacher model in conjunction with response-based knowledge. The feature-based knowledge transfer\cite{gou_knowledge_2020} can be generally formulated as,
        \begin{equation}
            L_{FesD}(f_t(x), f_s(x)) = \mathfrak{L}_{F}(\Phi_t(f_t(x)), \Phi_s(f_s(x))),
        \label{eq:4}
        \end{equation}
        where the feature maps of the intermediate layers of the teacher and student models are denoted as $f_t(x)$ and $f_s(x)$, respectively. The utilisation of the transformation function $\Phi_t(f_t(x))$ and $\Phi_s(f_s(x))$ is commonly used in cases where the feature maps of the teacher and student models exhibit different shapes. The $\mathfrak{L}_{F}()$ refers to the similarity function employed to match the feature maps of both the teacher and student models.
        
    \subsubsection{Relation-based Knowledge in Teacher Models}
        The relation-based knowledge explores the associations between different feature maps obtained through different layers of the network/s.

\begin{figure}[htbp]
                \centering
                	\includegraphics[width=1.0\linewidth]{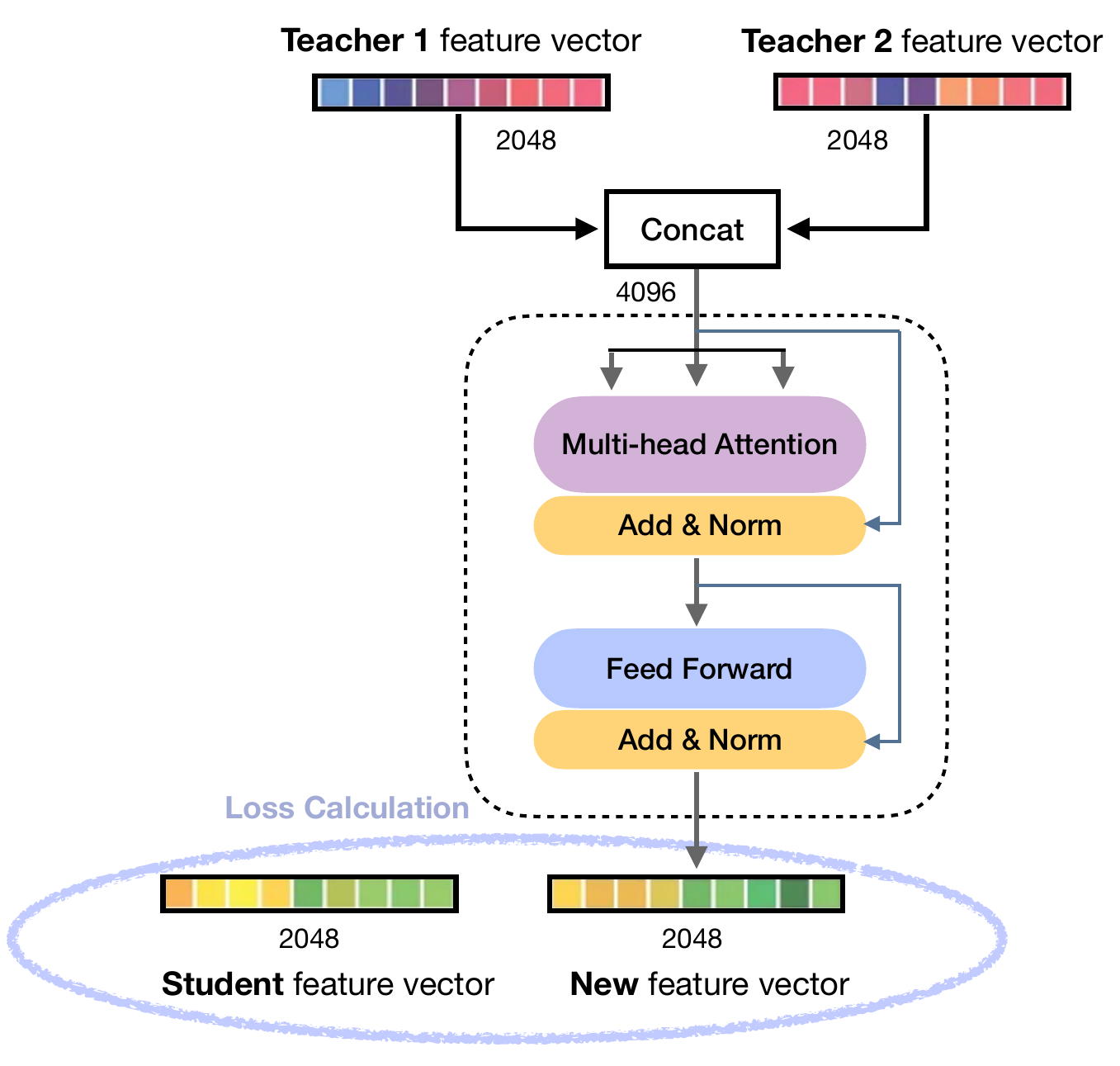}
        	\caption{Our relation-based Knowledge transfer mechanism: Multi-head attention-based fusion for generating new feature vector to guide Student learning.}
        	\label{fig:relational_kd}
        \end{figure}

        The main purpose of our relation-based approach is to support knowledge transfer through multiple teachers. In order to achieve this we utilise the feature maps from multiple teachers and generate a new single informative feature map ($f_{t*}$) for the student to learn from. We formulate our relation-based knowledge transfer by utilising the multi-head attention mechanism used in the Transformer networks introduced in \cite{vaswani_attention_2017}. We adopt a similar functionality to the encoder module of the Transformer network to fuse the important salient features of teacher networks to generate a more informative feature map that is able to guide the student model to achieve improved performance (illustrated in Fig. \ref{fig:relational_kd}). The original transformer network only utilise the multi-head attention techniques to map the correspondences between a single input stream at a time while we adopt this as a fusion mechanism to map the associations between the features passed from the two teachers. This jointly allows the network to learn a better feature representation that would be more capable of guiding the student network. 

        Once the new feature map $f_{t*}$ is generated through the multi-head attention block, the loss can be formulated as,

        \begin{equation}
            L_{RelD}(f_{t*}(x), f_s(x)) = \mathfrak{L}_{F}(\Phi_t(f_{t*}(x)), \Phi_s(f_s(x))),
        \end{equation}

        which is similar to Eq.\ref{eq:4}, except the feature map of the teacher ($f_{t*}$) is generated by utilising the feature maps of multiple teachers. This formulation merits the overall classification performance as the multi-head attentions are learned throughout the model training phase, such that it not only encode the feature map but also learns better feature maps that can support student model learning.


    In all three knowledge distillation methods, we follow offline distillation where the teacher models are pre-trained prior to applying them within the knowledge distillation framework. Considering the amount of data available for our classification task, we believe this is the most suitable setting as an online approach would require a large-scale dataset to train the teacher/s and student models concurrently.

\section{Experimental Setup}
    \subsection{Datasets}

    This study examined two publicly available datasets regarding the gastrointestinal tract, namely Kvasir-V2 \cite{pogorelov_kvasir_2017} and HyperKvasir \cite{borgli_hyperkvasir_2020}. The former is a balanced endoscopy dataset consisting of images published in 2017, while the latter is an imbalanced and comprehensive dataset comprising both images and videos posted in 2020 and is considered the largest of its kind. The data utilised in these studies was obtained from authentic gastro- and colonoscopy examinations conducted at Baerum Hospital in Norway. Furthermore, a portion of the data was annotated by experienced gastrointestinal endoscopists. The relevant ethical approvals for data collection have been obtained by the dataset authors. 
    
    \subsubsection{KVASIR-V2}
        The KVASIR-V2 dataset is composed of endoscopic images depicting various GI-tract-based diseases and anatomical landmarks \cite{pogorelov_kvasir_2017}. The dataset encompasses 7,200 high-quality RGB images showing diverse gastrointestinal diseases such as polyps, ulcerative colitis, pylorus, and esophagitis, as well as anatomical landmarks such as cecum and z-line. The KVASIR-V2 dataset demonstrates a satisfactory level of balance among its classes, ensuring adequate representation of each disease category. A balanced distribution of classes serves to mitigate the potential for bias towards particular classes, thereby enabling the training and evaluation of classification models with greater resilience.

        In our study, we made the decision to exclude the "Dyed Polyps" and "Dyed Resection Margins" categories from the KVASIR-V2 dataset. The staining process can result in unique visual features in dyed polyps and resection margins, which may not necessarily be indicative of the underlying pathology. The objective of eliminating these classes is to establish a more consistent and concentrated dataset, thereby enhancing the precision of disease classification. The KVASIR-V2 dataset has been randomly divided into training, validation, and testing datasets to ensure a reliable evaluation. The training dataset comprises 60\% (3,600 images out of 6,000 images by removing two 'dyed' classes) of the dataset, enabling the model to learn from a significant portion of the available data. The validation dataset includes 20\% (1200 images) of the dataset, serving as a separate set for tuning hyperparameters and tracking model performance during training. The remaining 20\% (1,200 images) of the dataset constitutes the testing dataset, which serves as an independent set to evaluate the final performance of the trained model.  
        

        Preprocessing techniques were implemented prior to model training to ensure data compatibility and optimise model performance. The preprocessing stage encompassed the resizing of the images to a standardised resolution of 224X224. The objective of the preprocessing procedures was to mitigate discrepancies and enhance the model's capacity to generalise across diverse samples.
    
    \subsubsection{Hyper-KVASIR}
        The dataset referred to as Hyper-KVASIR comprises 10,662 images annotated by medical professionals, serving as ground truth labels for disease classification and stored in the JPEG file format \cite{borgli_hyperkvasir_2020}. The dataset consists of 23 categories of gastrointestinal findings, encompassing a broad spectrum of pathologies and anatomical abnormalities. In addition, it displays class imbalance, a prevalent issue in the medical domain where certain outcomes are more prevalent than others. In order to ensure the reliability of our models, we executed suitable data splitting, whereby a segment of the dataset was assigned for the purpose of training (60\% of the datasets), validation (20\% of the datasets), and testing (20\% of the datasets). Similar Preprocessing techniques were utilised as in KVASIR-V2.

\subsection{Implementation Details}
\label{sec:imp_details}
        As the teacher models we choose three ResNet architectures \cite{he_deep_2015} ResNet50, ResNet101, and ResNet152, which are defined by 50, 101, and 152 layers, respectively. These ResNet architectures are widely used in image classification tasks \cite{alom_history_2018} as they are capable of learning features reasonably well. Our knowledge-based learning frameworks operate offline, hence we utilise the pre-trained ImageNet weights \cite{joshi_survey_2020} for all three teacher networks and follow two settings with and without fine-tuning the networks with the endoscopic image data. During the fine-tuning of the teacher models, we have fine-tuned all layers except the first 4 layers of the networks.
        
        Motivated by the ResNet architecture we design the lightweight student network, by adapting the layers 2D convolution, BatchNorm2D, Relu, max pooling, and adaptive average pooling layer followed by a flatten and fully connected layers respectively. This student network structure is further visualized in Fig. \ref{fig:framework}. The individual training (i.e. without KD) of teacher and student models was done using categorical cross-entropy loss and the Adam optimiser \cite{kingma_adam_2017} with a learning rate of 0.0001. For the KD-based learning through, we incorporated the Mean Squared Error for additional loss calculations.
        
\subsection{Model Complexities}

In Tab. \ref{tab:modelSize} we present the comparison of the teacher and the student model complexities based on the total parameters and the model sizes. It is seen that the model complexity reduction is extremely high, enabling the student model, with performance comparable to the teacher model, to be implemented in low computational resource platforms.  

\begin{table}[!h]
\centering
\caption{The model complexities of the student model and the teacher models. Here the total parameters and the model sizes of teachers are based on individual teacher models.}
\label{tab:modelSize}
\resizebox{0.95\columnwidth}{!}{%
\begin{tabular}{|l|r|r|}
\hline
\textbf{Model} & \textbf{\begin{tabular}[c]{@{}l@{}}Total Number\\ of Parameters\end{tabular}} & \textbf{Model Size (bytes)} \\ \hline
Student/KD Models & 51,895    & 210,823     \\ \hline
ResNet50      & 25,557,032 & 94,544,951  \\ \hline
ResNet101     & 44,549,160 & 170,830,157 \\ \hline
ResNet152     & 60,192,808 & 233,696,481 \\ \hline
\end{tabular}%
}
\end{table}

\section{Experimental Results and Discussion}
    The initial experiments were conducted on the teacher models (considering both pre-trained and fine-tuned weights) and our proposed lightweight student model evaluating them on the endoscopy datasets. Then these teacher models are utilised in different KD settings in order to achieve a better student model that is able to obtain enhanced performance compared to the performance of the student model learned without KD-based learning. We utilise the following notations to refer to different networks with different learning settings.     

\begin{itemize}
    \item \textbf{STU}: The lightweight student model(see Sec. \ref{sec:imp_details} for student model details) trained without KD-based learning.
    \item \textbf{RN50}: ResNet50 model using either original pre-trained weights or fine-tuned weights.  
    \item \textbf{RN101}: ResNet101 using either original pre-trained weights or fine-tuned weights. 
    \item \textbf{RN152}: ResNet152 using either original pre-trained weights or fine-tuned weights.
    \item \textbf{KD\_RESP\_RN50}: The student model trained through response-based KD with RN50 as the teacher.
    \item \textbf{KD\_RESP\_RN101}: The student model trained through response-based KD with RN101 as the teacher model.
    \item \textbf{KD\_RESP\_RN152}: The student model trained through response-based KD with RN152 as the teacher model.
    \item \textbf{KD\_FEAT\_RN50}: The student model trained through feature-based KD with RN50 as the teacher model.
    \item \textbf{KD\_FEAT\_RN101}: The student model trained through feature-based KD with RN101 as the teacher model.
    \item \textbf{KD\_FEAT\_RN152}: The student model trained through feature-based KD with RN152 as the teacher model.
    \item \textbf{KD\_FEAT\_T2\_1}: The student model trained through feature-based KD with two teachers (R50 and R101).
    \item \textbf{KD\_FEAT\_T2\_2}: The student model trained through feature-based KD with two teachers (R101 and R152).
    \item \textbf{KD\_FEAT\_T3}: The student model trained through feature-based KD with all three teachers.
    \item \textbf{KD\_REL\_T2}: The student model trained through relation-based KD with two teachers.
\end{itemize}

It should be noted that all response-based models and feature-based approaches rely on mean squared error calculations. Tab. \ref{tab:kvasirv2reaults} and Tab. \ref{tab:hyperkvasirv2reaults} include the experimental results on the datasets KVASIR-V2 and Hyper-KVASIR respectively. These tables include the overall classification performance considering all classes (including both diseases and anatomical landmarks) available through each dataset. It is apparent that the teacher models achieve the highest classification performance compared to the student models with or without KD. This is mainly due to the higher network capacity of the ResNet models allowing more space for learning the non-linearity within the data. However, as explained before the higher network capacity also results in a large number of parameters that requires training, hence the increased network size (see Tab. \ref{tab:modelSize}). In this work, the main objective is to utilize KD-based learning to achieve a lightweight model that is able to perform reasonably well. When the student model without KD-based learning (i.e. STU) is compared against the teacher models we can notice around a 10\%-12\% accuracy gap that makes the teacher models more suitable to be used in the KD-based learning of the student network. 

In KD-based learning, the first 3 evaluations were performed considering the response-based features. The 3 settings are formulated considering the 3 teachers selected in this study (i.e. RN50, RN101, and RN152). Among these, the RN152-based settings (i.e. KD\_RESP\_RN152) have achieved the highest classification performance for both datasets where the accuracy improvements are 1.75\% and 2.05\% for KVASIR-V2 and Hyper-KVASIR respectively. We believe this is mainly due to the fact that RN152 achieves higher performance compared to other teacher models, hence it has more knowledge that can be transferred to the student.

The remaining single teacher-based frameworks are the three feature-based learning settings (i.e. KD\_FEAT\_RN50, KD\_FEAT\_RN101, and KD\_FEAT\_RN152). As mentioned before, in this setting we utilise the hidden features from a later fully connected layer in addition to the response-based outputs during the KD-based training. By adapting the feature-based knowledge transfer we have noticed some improvements in results. For example, for KVASIR-V2, the feature-based approach with the teacher RN50 (KD\_FEAT\_RN50) has achieved a 0.26\% increase compared to the response-based approach with the same teacher model (KD\_RESP\_RN50). However, for the Hyper-KVASIR dataset, the same experiments showed an opposite behaviour where the accuracy was decreased by 0.23\%. For both datasets, the feature-based setting (KD\_FEAT\_RN101) has shown more than a 0.72\% accuracy increase compared to its response-based approach (i.e. KD\_RESP\_RN101). We believe this inconsistency is mainly due to the characteristics of the datasets and the nature of the features passed through different teacher models.

In multi-teacher-based knowledge transfer, we could not achieve considerable improvements through the 2 settings KD\_FEAT\_T2\_1 and KD\_FEAT\_T2\_2. However, the use of all three teachers has increased the accuracy by 0.42\% and 0.52\% for KVASIR-V2 and Hyper-KVASIR datasets respectively, compared to the STU-based accuracy. However, this is not sufficient as some single-teacher settings (both response-based and feature-based) are able to achieve higher improvements with respect to classification. We suspect that this is due to the simple mean square error-based mechanism we utilized for the knowledge transfer.

For the final experiment, we have adapted the multi-head attention introduced in \cite{vaswani_attention_2017} in order to fuse the important features from two teachers towards creating a more informative feature vector for the student to learn from. This has significantly increased the accuracy of the student from its STU setting (i.e. 3\% and 4.97\% for KVASIR-V2 and Hyper-KVASIR respectively). Such multi-head attention is often used to map the relations among the features within the same inputs. However, this study signifies the use of multi-head attention for the fusion of multi-features in order to create a more informative feature vector to support KD-based learning. Once the training is performed through the relation-based knowledge transfer the trained lightweight model alone can be executed in low resource environment (e.g. on a capsule endoscope). The model size comparisons in Tab. \ref{tab:modelSize} further highlight the suitability of our lightweight student model is approximately 448 times smaller than the lightest ResNet model (i.e. ResNet50).

\begin{table*}[!h]
\centering
\caption{Evaluation Results on KVASIR-V2 Dataset}
\label{tab:kvasirv2reaults}
\resizebox{0.8\textwidth}{!}{%
\begin{tabular}{|c|c|c|c|c|c|c|}
\hline
\textbf{Type}                                                                                   & \textbf{Model}                    & {\color[HTML]{656565} \textbf{Fine-Tuning}} & {\color[HTML]{010066} \textbf{Accuracy}} & {\color[HTML]{010066} \textbf{Precision}} & {\color[HTML]{010066} \textbf{F1 Score}} & {\color[HTML]{010066} \textbf{Recall}} \\ \hline
                                                                                    &                                   & Fine-tuned                                  & 90.6\%                                  & 90.9\%                                   & 90.3\%                                  & 90.3\%                                \\ \cline{3-7} 
                                                                                    & \multirow{-2}{*}{RN50}            & Pre-trained                                 & 85.6\%                                  & 85.9\%                                   & 85.8\%                                  & 85.8\%                                \\ \cline{2-7} 
                                                                                    &                                   & Fine-tuned                                  & 91.8\%                                  & 92.4\%                                   & 92.3\%                                  & 92.3\%                                \\ \cline{3-7} 
                                                                                    & \multirow{-2}{*}{RN101}           & Pre-trained                                 & 86.5\%                                  & 86.1\%                                   & 85.9\%                                  & 85.9\%                                \\ \cline{2-7} 
                                                                                    &                                   & Fine-tuned                                  & 92.6\%                                  & 92.6\%                                   & 92.4\%                                  & 92.4\%                                \\ \cline{3-7} 
\multirow{-6}{*}{\textbf{\begin{tabular}[c]{@{}l@{}}Teacher\\ Models\end{tabular}}} & \multirow{-2}{*}{RN152}           & Pre-trained                                 & 85.2\%                                  & 86.1\%                                   & 85.9\%                                  & 85.9\%                                \\ \hline
\textbf{\begin{tabular}[c]{@{}l@{}}Student\\ Model\end{tabular}}                    & STU                               & -                                           & 80.6\%                                  & 80.7\%                                   & 80.8\%                                  & 80.8\%                                \\ \hline
                                                                                    &                                   & Fine-tuned                                  & 81.4\%                                  & 81.7\%                                   & 81.4\%                                  & 81.4\%                                \\ \cline{3-7} 
                                                                                    & \multirow{-2}{*}{KD\_RESP\_RN50}  & Pre-trained                                 & 81.8\%                                  & 81.3\%                                   & 81.1\%                                  & 81.3\%                                \\ \cline{2-7} 
                                                                                    &                                   & Fine-tuned                                  & 82.17\%                                  & 82.46\%                                   & 82.17\%                                  & 82.17\%                                \\ \cline{3-7} 
                                                                                    & \multirow{-2}{*}{KD\_RESP\_RN101} & Pre-trained                                 & 81.0\%                                  & 81.1\%                                   & 81.0\%                                  & 81.0\%                                \\ \cline{2-7} 
                                                                                    &                                   & Fine-tuned                                  & 82.3\%                                  & 81.9\%                                   & 81.9\%                                  & 81.9\%                                \\ \cline{3-7} 
                                                                                    & \multirow{-2}{*}{KD\_RESP\_RN152} & Pre-trained                                 & 81.1\%                                  & 81.1\%                                   & 81.3\%                                  & 81.3\%                                \\ \cline{2-7} 
                                                                                    & KD\_FEAT\_RN50                    & -                                           & 82.0\%                                  & 82.1\%                                   & 82.0\%                                  & 82.0\%                                \\ \cline{2-7} 
                                                                                    & KD\_FEAT\_RN101                   & -                                           & 82.7\%                                  & 82.2\%                                   & 82.3\%                                  & 82.3\%                                \\ \cline{2-7} 
                                                                                    & KD\_FEAT\_RN152                   & -                                           & 81.9\%                                  & 81.7\%                                   & 82.0\%                                  & 82.0\%                                \\ \cline{2-7} 
                                                                                    & KD\_FEAT\_T2\_1                   & -                                           & 80.9\%                                  & 81.0\%                                   & 80.8\%                                  & 80.8\%                                \\ \cline{2-7} 
                                                                                    & KD\_FEAT\_T2\_2                   & -                                           & 79.3\%                                  & 79.3\%                                   & 79.3\%                                  & 79.3\%                                \\ \cline{2-7} 
                                                                                    & KD\_FEAT\_T3                      & -                                           & 81.0\%                                  & 81.4\%                                   & 81.3\%                                  & 81.3\%                                \\ \cline{2-7} 
\multirow{-13}{*}{\textbf{\begin{tabular}[c]{@{}l@{}}KD\\ Models\end{tabular}}}     & KD\_REL\_T2                       & -                                           & 83.6\%                                  & 83.4\%                                   & 83.2\%                                  & 83.2\%                                \\ \hline
\end{tabular}%
}
\end{table*}

\begin{table*}[!h]
\centering
\caption{Evaluation Results on Hyper-KVASIR Dataset.}
\label{tab:hyperkvasirv2reaults}
\resizebox{0.8\textwidth}{!}{%
\begin{tabular}{|c|c|c|c|c|c|c|}
\hline
\textbf{Type}                                                                 & \textbf{Model}                    & {\color[HTML]{656565} \textbf{Fine-Tuning}} & {\color[HTML]{010066} \textbf{Accuracy}} & {\color[HTML]{010066} \textbf{Precision}} & {\color[HTML]{010066} \textbf{F1 Score}} & {\color[HTML]{010066} \textbf{Recall}} \\ \hline
                                                                           &                                   & Fine-tuned                                  & 88.5\%                                  & 87.9\%                                   & 88.7\%                                  & 88.7\%                                \\ \cline{3-7} 
                                                                           & \multirow{-2}{*}{RN50}            & Pre-trained                                 & 83.9\%                                  & 82.7\%                                   & 84.3\%                                  & 84.3\%                                \\ \cline{2-7} 
                                                                           &                                   & Fine-tuned                                  & 88.8\%                                  & 87.3\%                                   & 88.9\%                                  & 88.9\%                                \\ \cline{3-7} 
                                                                           & \multirow{-2}{*}{RN101}           & Pre-trained                                 & 83.5\%                                  & 80.9\%                                   & 83.0\%                                  & 83.0\%                                \\ \cline{2-7} 
                                                                           &                                   & Fine-tuned                                  & 89.3\%                                  & 88.5\%                                   & 89.5\%                                  & 89.5\%                                \\ \cline{3-7} 
\multirow{-6}{*}{\textbf{\begin{tabular}[c]{@{}l@{}}Teacher\\ Models\end{tabular}}} & \multirow{-2}{*}{RN152}           & Pre-trained                                 & 82.5\%                                  & 80.2\%                                   & 82.7\%                                  & 82.7\%                                \\ \hline
\textbf{\begin{tabular}[c]{@{}l@{}}Student\\ Model\end{tabular}}                    & STU                               & -                                           & 77.0\%                                  & 75.4\%                                   & 77.1\%                                  & 77.1\%                                \\ \hline
                                                                           &                                   & Fine-tuned                                  & 79.0\%                                  & 76.6\%                                   & 79.0\%                                  & 79.0\%                                \\ \cline{3-7} 
                                                                           & \multirow{-2}{*}{KD\_RESP\_RN50}  & Pre-trained                                 & 78.6\%                                  & 76.1\%                                   & 78.2\%                                  & 78.2\%                                \\ \cline{2-7} 
                                                                           &                                   & Fine-tuned                                  & 78.1\%                                  & 76.3\%                                   & 78.1\%                                  & 78.1\%                                \\ \cline{3-7} 
                                                                           & \multirow{-2}{*}{KD\_RESP\_RN101} & Pre-trained                                 & 78.6\%                                  & 76.3\%                                   & 78.5\%                                  & 78.5\%                                \\ \cline{2-7} 
                                                                           &                                   & Fine-tuned                                  & 79.1\%                                  & 77.1\%                                   & 79.1\%                                  & 79.1\%                                \\ \cline{3-7} 
                                                                           & \multirow{-2}{*}{KD\_RESP\_RN152} & Pre-trained                                 & 78.7\%                                  & 76.1\%                                   & 78.3\%                                  & 78.3\%                                \\ \cline{2-7} 
                                                                           & KD\_FEAT\_RN50                    & -                                           & 78.8\%                                  & 75.4\%                                   & 78.8\%                                  & 78.8\%                                \\ \cline{2-7} 
                                                                           & KD\_FEAT\_RN101                   & -                                           & 79.5\%                                  & 76.4\%                                   & 79.0\%                                  & 79.0\%                                \\ \cline{2-7} 
                                                                           & KD\_FEAT\_RN152                   & -                                           & 78.6\%                                  & 76.0\%                                   & 78.6\%                                  & 78.6\%                                \\ \cline{2-7} 
                                                                           & KD\_FEAT\_T2\_1                   & -                                           & 77.5\%                                  & 75.1\%                                   & 77.5\%                                  & 77.5\%                                \\ \cline{2-7} 
                                                                           & KD\_FEAT\_T2\_2                   & -                                           & 78.3\%                                  & 76.2\%                                   & 78.3\%                                  & 78.3\%                                \\ \cline{2-7} 
                                                                           & KD\_FEAT\_T3                      & -                                           & 77.6\%                                  & 75.2\%                                   & 77.6\%                                  & 77.6\%                                \\ \cline{2-7} 
\multirow{-13}{*}{\textbf{\begin{tabular}[c]{@{}l@{}}KD\\ Models\end{tabular}}}     & KD\_REL\_T2                       & -                                           & 82.0\%                                  & 81.8\%                                   & 82.0\%                                  & 82.0\%                                \\ \hline
\end{tabular}%
}
\end{table*}

\begin{figure*}[!h]
     \centering
     \begin{subfigure}[b]{0.45\textwidth}
         \centering
         \includegraphics[width=\textwidth]{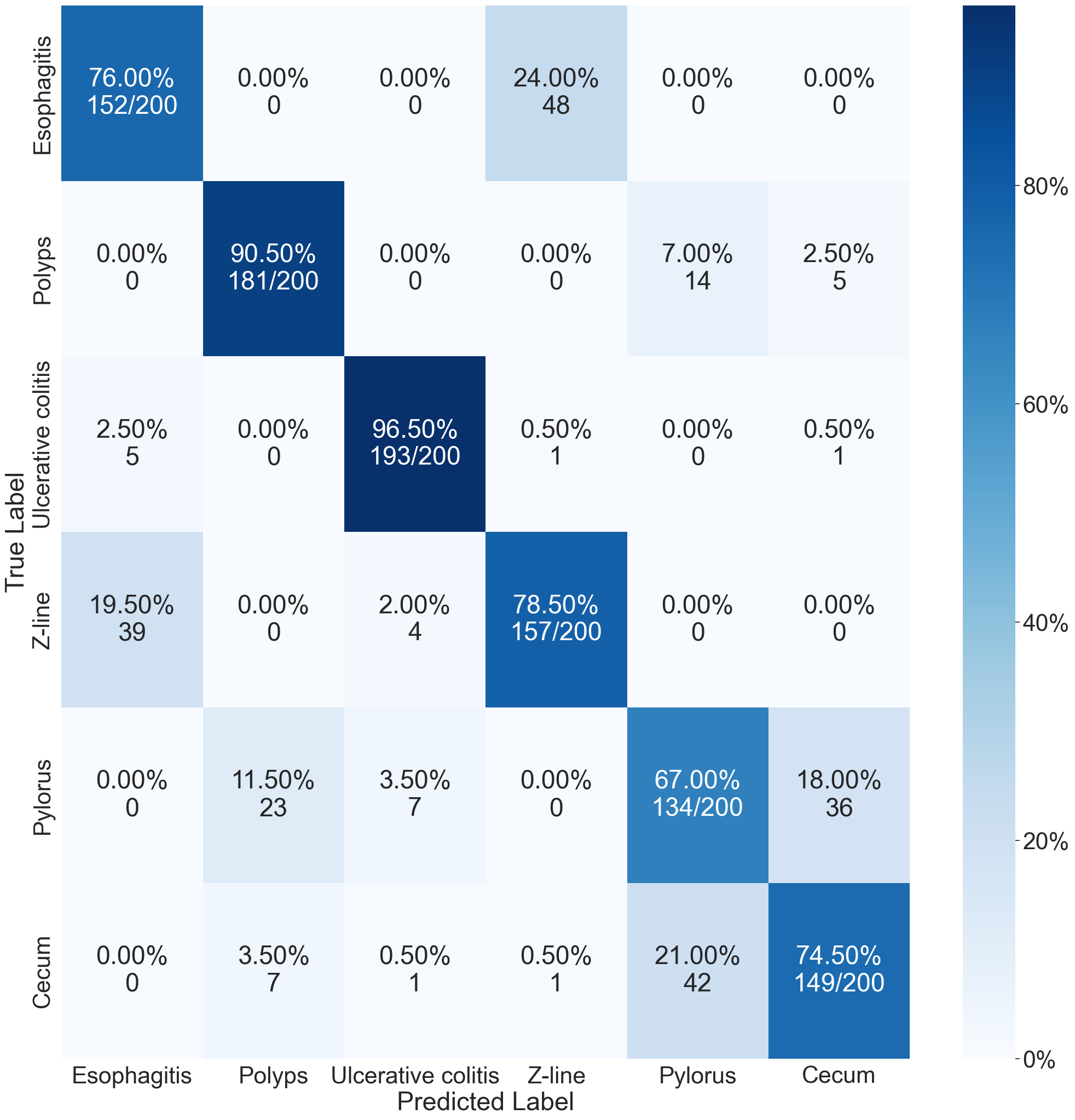}
         \caption{Student only (STU)}
         \label{fig:stu_cm}
     \end{subfigure}
     \hfill
     \begin{subfigure}[b]{0.45\textwidth}
         \centering
         \includegraphics[width=\textwidth]{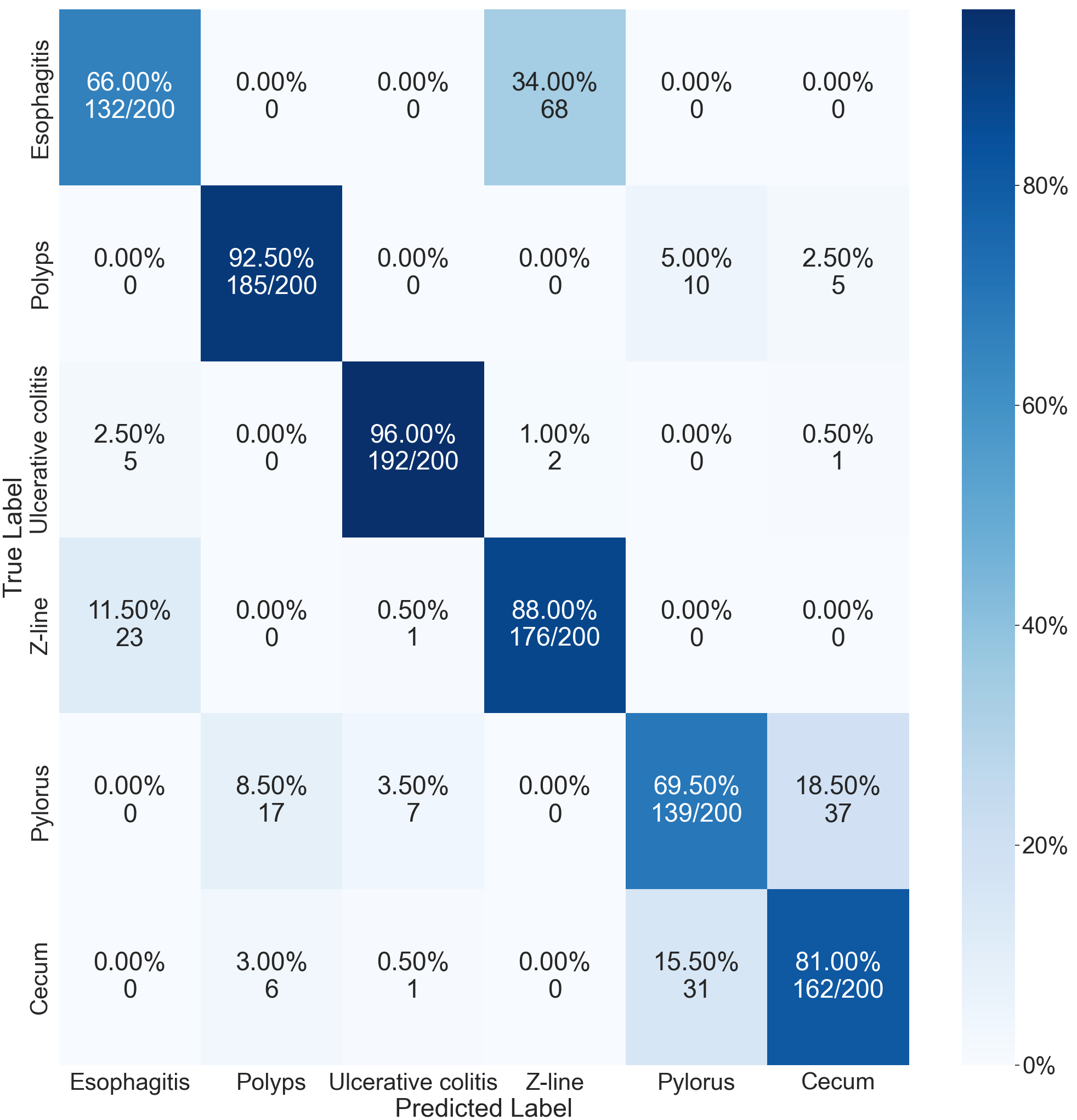}
         \caption{Single teacher KD-based learning (KD\_FEAT\_RN101)}
         \label{fig:kd_cm}
     \end{subfigure}
     \hfill
        \caption{Confusion matrix visualisations of student model with and without KD-based learning.}
        \label{fig:confusion_mats}
\end{figure*}

In order to illustrate the benefit of using single-teacher KD-based learning, we visualize the confusion matrices for the two settings, STU and KD\_FEAT\_RN101 in Fig. \ref{fig:confusion_mats}. Even though we noticed a considerable degradation in the class-based classification accuracies for Esophagitis (i.e. 10.0\%) and Polyps (i.e. 0.17\%) in KD\_FEAT\_RN101 setting compared to the student-only STU setting, the class-based accuracies for all remaining classes (i.e. Polyp, Z-line, Pylorus, and Cecum) has significantly improved. We especially noticed 9.5\% and 6.5\% accuracy increases for the classes Z-line and Cecum. In the STU setting, we have noticed around 21\% of Cecum images have been misclassified as Pylorus. However, through KD-based learning, this has been reduced to 15.5\%.            

\begin{figure*}[!h]
     \centering
     \begin{subfigure}[b]{0.49\textwidth}
         \centering
         \includegraphics[width=\textwidth]{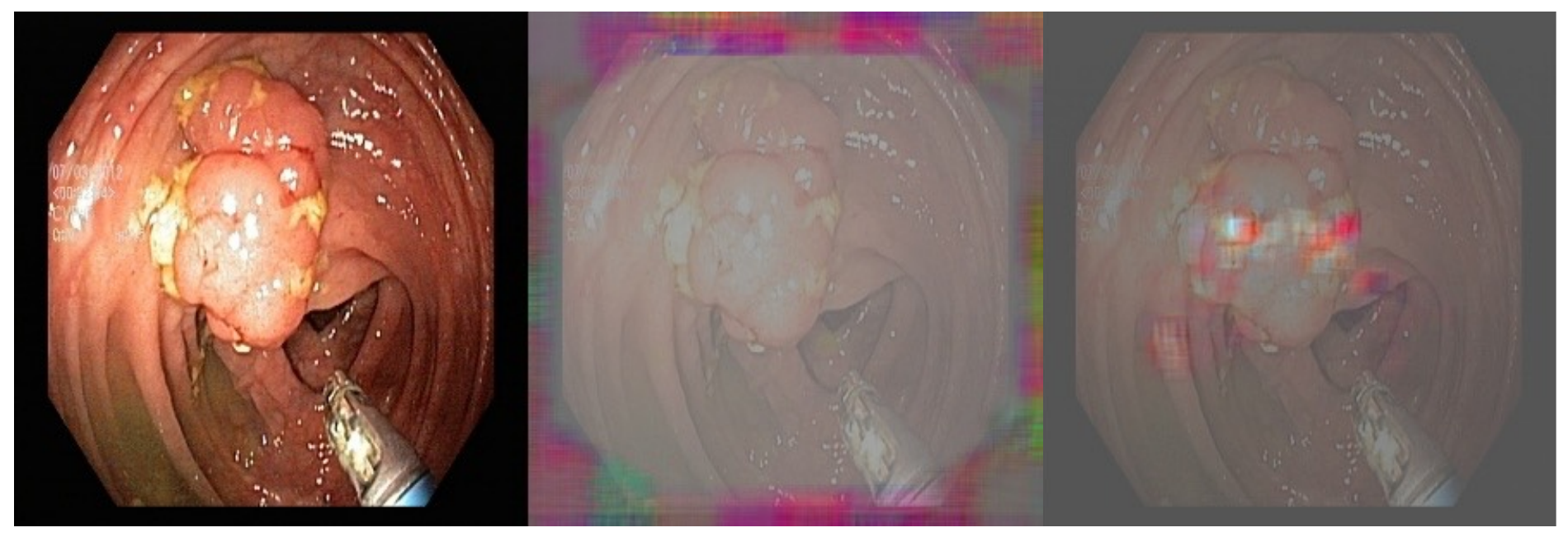}
         \caption{Polyp}
         \label{fig:gradcam1}
     \end{subfigure}
     \hfill
     \begin{subfigure}[b]{0.49\textwidth}
         \centering
         \includegraphics[width=\textwidth]{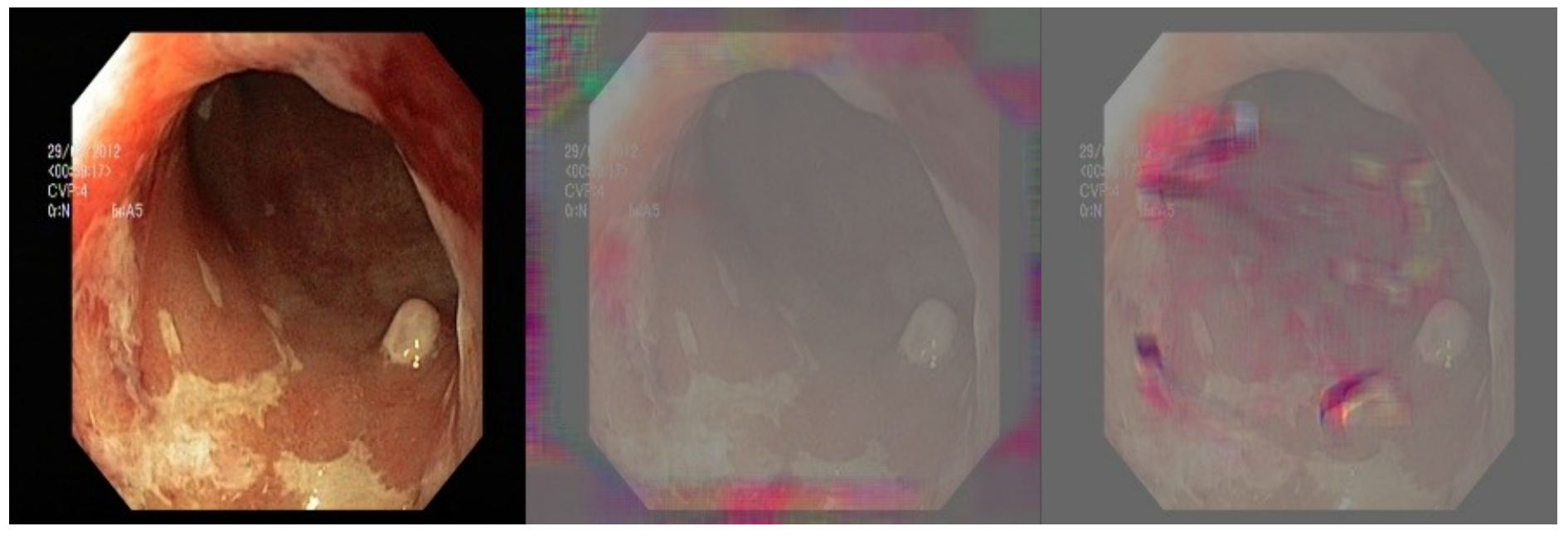}
         \caption{Ulcerative Colitis}
         \label{fig:gradcam2}
     \end{subfigure}
     \hfill
     \begin{subfigure}[b]{0.49\textwidth}
         \centering
         \includegraphics[width=\textwidth]{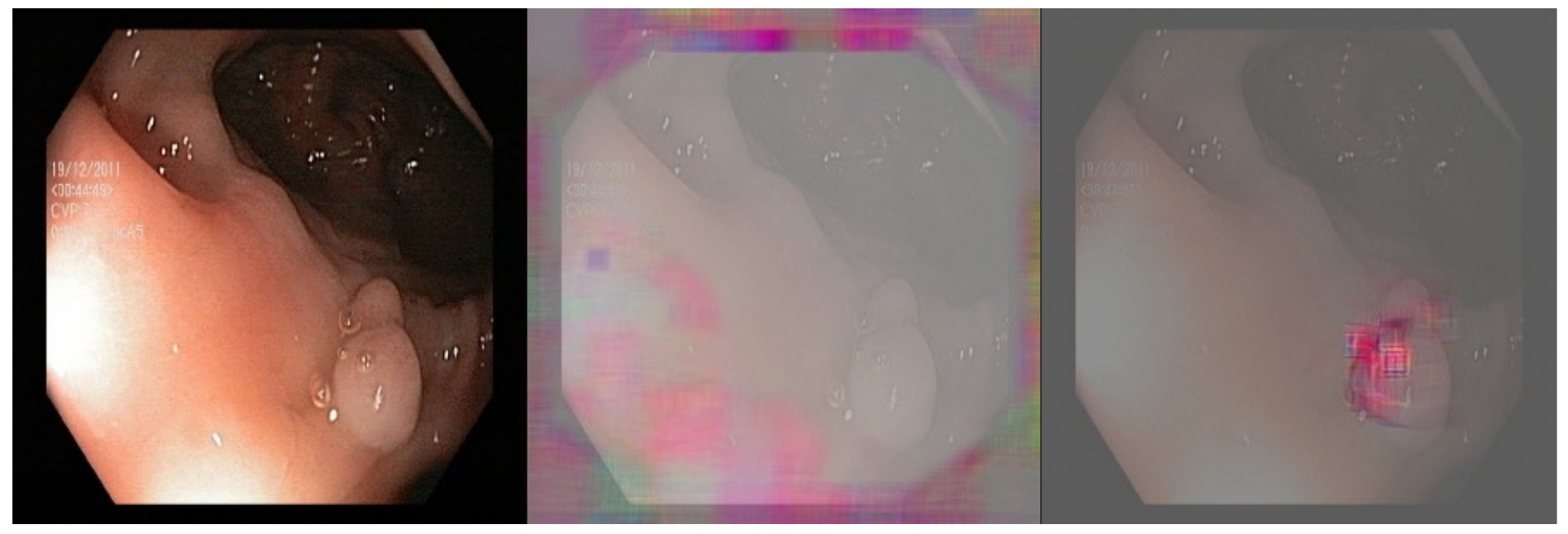}
         \caption{Polyp}
         \label{fig:gradcam3}
     \end{subfigure}
     \begin{subfigure}[b]{0.49\textwidth}
         \centering
         \includegraphics[width=\textwidth]{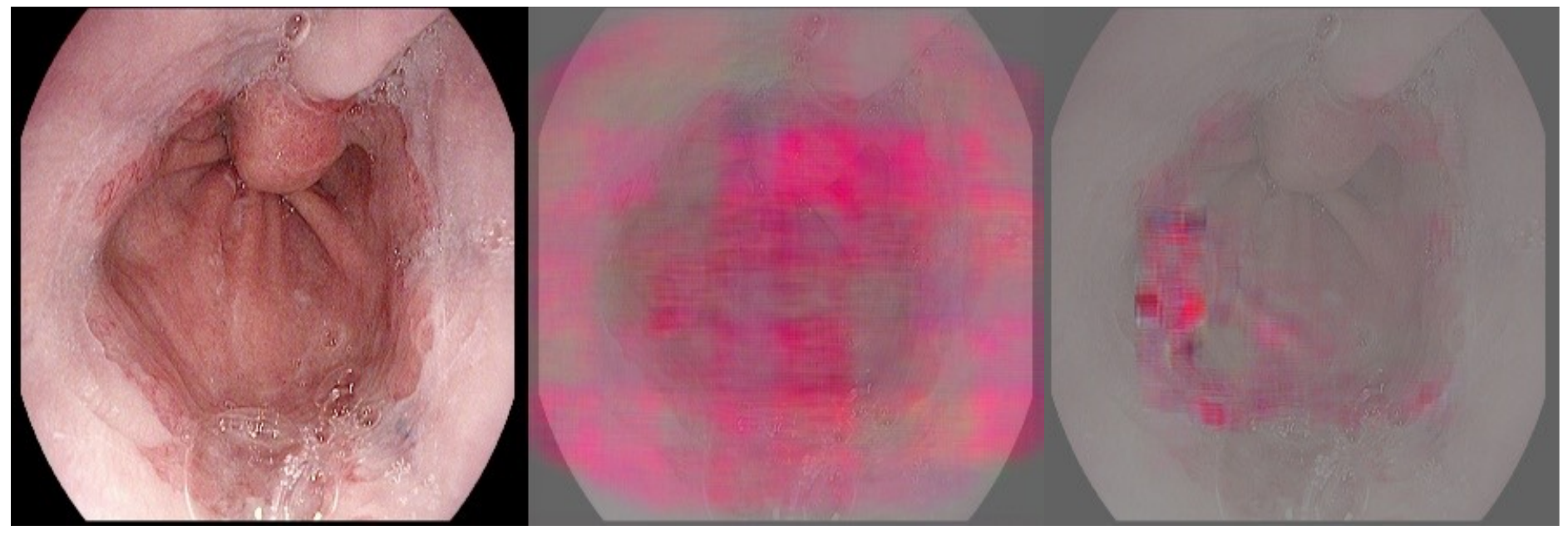}
         \caption{Esophagitis}
         \label{fig:gradcam4}
     \end{subfigure}
        \caption{Model interpretations through GradCam visualisations. In each sub-figure, the original endoscopy image(left), student-only model interpretations (middle), and KD\_REL\_T2 based interpretations (right) are visualised. Note that the brighter colours (i.e. red and pink) show higher attention.}
        \label{fig:visualisations}
\end{figure*}

In order to illustrate how the introduced relation-based approach (KD\_REL\_T2) has complemented student learning, we visualize the model interpretations on a few endoscopic images with diseases (such as esophagitis, polyps, and ulcerative colitis) in the KVASIR dataset (see Fig. \ref{fig:visualisations}). These illustrations clearly highlight how attention is improved through KD-based learning. For example, in Fig. \ref{fig:gradcam1}, the student-only model (i.e. STU) focus is mainly distributed towards the boundaries of the image which could not capture any polyp-related information. However, through KD-based learning (i.e. KD\_REL\_T2) this focus has moved to the areas on the polyp which aided in better classification performance. The remaining instances of esophagitis and ulcerative colitis also illustrate similar behaviour.



\section{Conclusion}
The state-of-the-art endoscopy image classification methods are often based on deep neural networks that often require large amounts of resources to perform training. Considering this limitation of the existing methods which makes them unsuitable to run on low-resource environments (e.g. mobile applications) to obtain real-time performance, we investigated the use of KD-based learning to enhance the performance of a custom lightweight classification model. We have obtained promising outcomes with KD-based learning through single and multi-teacher-based settings. We further contribute a novel relation-based knowledge transfer mechanism that utilises two teachers' knowledge to generate a single feature vector that aided in guiding the student model learning. This has shown significant improvements in the classification performance of the student model on two endoscopic image datasets. Our findings further highlighted the relevance of utilizing multi-head attention-based fusion in generating more informative feature vectors to guide student learning. Our novel KD-based learning strategies enable the lightweight student network to achieve performance comparable to the heavy-weight teacher network, thereby enabling effective endoscopy image classification to be implemented in practical computing platforms typically present in hospital settings.


%



\section*{Acknowledgment}

The research presented in this paper was supported partly by the Queensland University of Technology (QUT) Center for Biomedical Technologies.

\ifCLASSOPTIONcaptionsoff
  \newpage
\fi



%
\printbibliography

%








\end{document}